\documentclass[10pt,twocolumn,letterpaper]{article}

\usepackage{cvpr}              
\definecolor{cvprblue}{rgb}{0.21,0.49,0.74}
\usepackage[pagebackref,breaklinks,colorlinks,allcolors=cvprblue]{hyperref}

\usepackage{booktabs}       
\usepackage{amsfonts}       
\usepackage{nicefrac}       
\usepackage{microtype}      
\usepackage{xcolor}         
\usepackage{float}         
\usepackage{graphicx}
\usepackage{enumitem}
\usepackage{wrapfig}
\usepackage{amsmath}
\usepackage{multirow}
\usepackage{tabularx}
\usepackage{adjustbox}
\usepackage{enumitem}
\usepackage{natbib}
\usepackage{stfloats}
\usepackage{pbalance}

\def\paperID{} 
\def\confName{CVPR}
\def\confYear{2026}

\title{THEval. Evaluation Framework for Talking Head Video Generation}

\author{
Nabyl Quignon\textsuperscript{1} \quad
Baptiste Chopin\textsuperscript{3} \quad
Yaohui Wang\textsuperscript{2}\textsuperscript{$\dagger$} \quad
Antitza Dantcheva\textsuperscript{1}
\\[6pt]
\textsuperscript{1}Inria Centre at Université Côte d'Azur
\quad\quad
\textsuperscript{2}Shanghai AI Laboratory
\quad\quad\quad
\\\textsuperscript{3}da/sec - Biometrics and Security Research Group, Hochschule Darmstadt
\quad\quad
}

\begin{document}
\twocolumn[{%
\renewcommand\twocolumn[1][]{#1}%
\maketitle

\includegraphics[width=\textwidth]{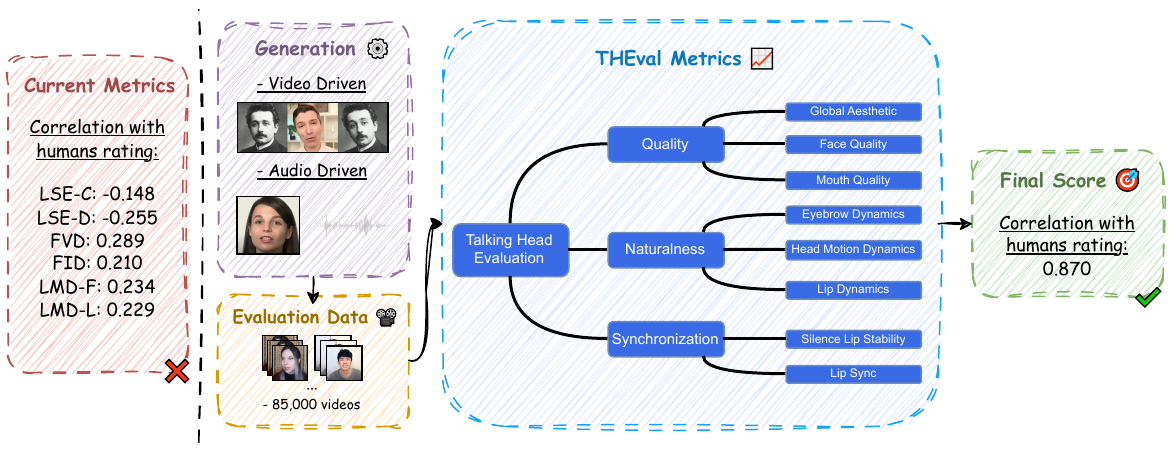}
\captionof{figure}{\textbf{Overview of the THEval benchmark.} We have generated talking head videos by 17 state-of-the-art video- and audio-driven methods, based on a dataset of over 5,000 videos spanning, resulting in 85,000 videos. We conduct a user study which demonstrates poor alignment between existing metrics (left red box) and human ratings. Motivated by this, we proceed to introduce the evaluation framework THEval, including 8 metrics related to (i) \textit{quality}, (ii) \textit{naturalness}, and (iii) \textit{synchronization} (center). These metrics are combined into a final score (right green box) that showcases a high correlation of 0.870 with human ratings, thereby constituting a new benchmark for evaluation of talking head videos. Code, dataset and leaderboard are available at \url{https://newbyl.github.io/theval_project_page/}}
\label{fig:teaser}
\vspace{0.5cm}
}]

\footnotetext[2]{Corresponding author.}

\begin{abstract}
Video generation has achieved remarkable progress, with generated videos increasingly resembling real ones. However, the rapid advancement in generation has outpaced the development of adequate evaluation metrics. Currently, the assessment of talking head generation primarily relies on limited metrics, evaluating general video quality, lip synchronization, and on conducting user studies. Motivated by this, we propose a \textit{new evaluation framework} comprising \textit{8 metrics} related to three dimensions (i) \textit{quality}, (ii) \textit{naturalness}, and (iii) \textit{synchronization}. In selecting the metrics, we place emphasis on efficiency, as well as alignment with human preferences. Based on these considerations, we aim to analyze fine-grained dynamics of head, mouth, and eyebrows, as well as face quality. Our extensive experiments on 85,000 videos generated by 17 state-of-the-art models suggest that while many algorithms excel in lip synchronization, they face challenges with generating expressiveness and artifact-free details. 
These videos were generated based on a novel real dataset that we have curated, in order to mitigate bias of training data. 
Our proposed benchmark framework is aimed at \textit{evaluating the improvement of generative methods focused on talking heads}.
\end{abstract}    
\section{Introduction}
\label{sec:intro}

Generative models have witnessed remarkable progress and are currently able to generate high-resolution, highly realistic images \citep{xu2024ufogen,xue2024raphael}  and videos \citep{wang2024microcinema,wang2024recipe}. However, the rapid advancement in generation has outpaced the development of adequate evaluation metrics, see details on existing metrics and their limitations in the Supplementary Material (SM) \ref{glossary}. 
Prominent metrics in image quality evaluation, such as the Fréchet Inception Distance (FID),  Inception Score (IS), and Learned Perceptual Image Patch Similarity (LPIPS) are limited in capturing quality and realism in generated data. 
Despite their limitations, 
most recent works in \textit{image} generation are  predominantly evaluated by the means of \textit{FID and IS} \citep{xu2024ufogen,yang2024improving}, as well as \textit{human user studies} \citep{shi2024instantbooth,chen2024subject}. 

\textbf{Video quality evaluation} 
encounters similar challenges as image quality assessment, primarily relying on metrics such as FID, Fréchet Video Distance (FVD), and Inception Score (IS) \citep{wang2024recipe,qing2024hierarchical}, which have limitations for high-quality videos. These metrics often omit motion quality and temporal coherence, necessitating human evaluation for such aspects. Alternative metrics  \citep{liu2024evalcrafter,huang2023vbench} have been proposed for \textit{general} video quality evaluation, assessing imaging and aesthetic quality, as well as for background consistency and motion smoothness in the context of temporal evaluation.

Deviating from general text-based video generation, \textit{talking head (TH) generation} involves an audio-speech sequence, employed to animate a face image. 
Evaluating this specific setting requires assessment of global image quality, as well as the quality of facial motions, including lip synchronization, facial expression, and head pose movement, all contributing to the video's overall naturalness. Currently, evaluations \citep{xu2024vasa,chen2024echomimic} focus on two main parameters: \textit{image quality}, measured by metrics including FID, FVD, SSIM, and PSNR, which share issues found in other video generation tasks, as well as \textit{lip synchronization quality}, assessed using Euclidean distances between landmarks or scores from the pretrained Syncnet neural network \citep{Chung16a}. However, studies have shown that Syncnet is unstable \citep{yaman2024audio} and sensitive to factors such as mouth cropping and head pose. Additionally, the commonly used Syncnet confidence score (LSE-C) and distance (LSE-D) have been found to correlate poorly with human preferences \citep{10647543}.

Motivated by the above and towards addressing the challenges of evaluating TH videos, we propose a new framework, referred to as \textsc{THEval}.  
Our main contributions include the following.

\begin{itemize}[left=5pt,noitemsep,nolistsep]
    \item We introduce \textsc{THEval}, a new framework with 8 fine-grained metrics across three dimensions (i) quality, (ii) naturalness, and (iii) synchronization that shows a Spearman correlation coefficient of  $\rho=0.87$ between our \textit{Final Score} and human ratings.
    \item We release a new, challenging evaluation dataset of over 5000 videos designed to test model generalization on unseen videos.
    \item Through an extensive benchmark of 17 SOTA audio- and video-driven models we provide a detailed analysis of related strengths and weaknesses.
    \item We conduct a user study, showcasing that THEval strongly correlates with human preferences. It also reveals that current metrics do not correlate well with user preferences. \\
\end{itemize}
 
Ultimately, we here make the case that existing metrics for evaluating TH-videos are highly limited and proceed to propose a novel evaluation framework THEval, see Figure \ref{fig:teaser}, which aligns with human preferences. THEval is intended for researchers and practitioners in the field of TH generation, providing a tool that identifies remaining challenges and fosters progress.
Towards this, we will make our benchmark publicly available, including dataset, code, and a live leaderboard. 
\section{Related Work}
\label{related_work}

\paragraph{Talking head video generation.} TH video generation can be \textit{video} or \textit{audio}-driven. Video-driven methods animate a face image using a driving video, effectively replacing the identity in the original footage. Early techniques utilized generative adversarial networks (GAN)-inversion or motion flow, while recent approaches focus on directly controlling the latent space \citep{Ni_2023_CVPR,wang2024lia}. Such methods tend to outperform audio-driven methods, as they reconstruct the motion and therefore encompass fewer degrees of freedom. 
In contrast, audio-driven methods entail fewer constraints in their effort to reproduce lip motion, expression, and head pose corresponding to the audio input. \textit{Text}-driven models  \citep{wang2022anyonenet,li2021write} commonly convert text to audio or phonemes before generating video, and therefore fall in this context under the category of 
audio-driven methods.

In this paper, we focus on audio- and video-driven TH generation that animate face images in RGB space \citep{Zhou2021Pose,Wang_2023_CVPR}, employing 
landmarks \citep{wang2021audio2head,gururani2022SPACE}, or face mesh representations \citep{oneshottalking,thies2020nvp}. While some methods produce 3D meshes, we note that THEval focuses solely on RGB videos. Audio-driven methods can be based on CNNs, LSTMs, or more recent diffusion models \citep{taklingheadgen,yu2023talking,shen2023difftalk}. Challenges in generation of TH have to do with 
accurate lip synchronization, as well as realistic face appearance, expressions and head movements. While state-of-the-art methods improve in generating related videos, evaluating such aspects remains an open challenge.

\paragraph{Image quality based metrics.} 
Currently image quality is evaluated with metrics such as FID, SSIM, PSNR \citep{xu2024vasa,chen2024echomimic}. %
FID compares the probability density distributions of real and generated images, SSIM measures the similarity between real and generated images, and PSNR evaluates noise levels in the output videos.
Such metrics face challenges with complex generated videos and are impacted by factors that do not affect video quality.
\begin{figure}
    \centering
    \includegraphics[width=0.8\columnwidth, keepaspectratio]{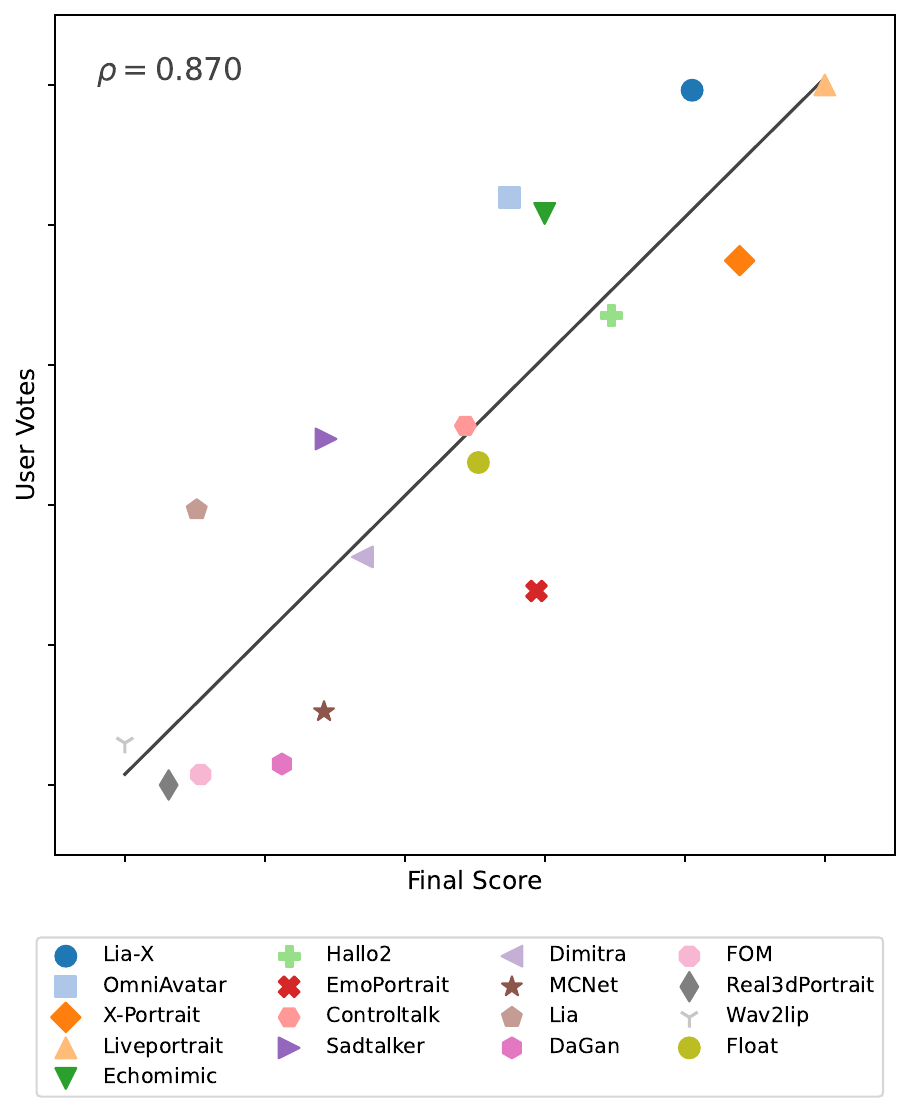}
    \caption{\textbf{THEval--Human Correlation.} A high Spearman correlation coefficient ($\rho = 0.870$) confirms THEval's strong alignment with human ratings. Each point represents a human preference for a state-of-the-art model win rate (y-axis) versus its THEval score (x-axis). This validation enables THEval to serve as an efficient proxy for costly user studies.}
    \label{fig:corr}
    \vspace{-5mm}
\end{figure}

Moreover, these metrics often do not align with human ratings for high-quality images and videos. In the context of audio-driven TH generation, experiments typically use small subsets of videos (a few hundred to a few thousand) in inference for efficiency. However, metrics such as FID (and by extension FVD) can be biased when assessed on limited samples, which may not provide a sufficient basis for generalization.

\paragraph{Facial landmark based metrics.} The LMD-F and LMD-M metrics assess facial expression and head movement using facial landmarks, showing better correlation with human evaluations than other metrics \citep{10647543}. However, they impart significant limitations. LMD-M penalizes small temporal lags that human evaluators might not notice, whereas LMD-F imposes strong penalties for differences in head motion and expression between generated videos and ground truth. 
This penalization is not justified as head motion and facial expression exhibit only weak correlation with audio sequences. We note that naturalness yields high human ratings, even in the presence of such discrepancies. Nevertheless, LMD metrics remain widely adopted, and the LMD-F variant is sometimes employed in evaluations involving real head poses, where it often produces favorable outcomes 
\citep{ma2023styletalk,ma2023dreamtalk}.

\paragraph{Syncnet metrics.} The most prominent metrics for lip synchronization are the Syncnet distance (LSE-D) and confidence score (LSE-C). Syncnet, a CNN-based network, aims to capture the correlation between audio and spatio-temporal features of the mouth region, calculating the audio offset (the number of frames by which audio is early or late compared to video). 
LSE-D represents the feature distance at the predicted offset, whereas LSE-C measures the difference between the minimum and median distances across various offsets. Syncnet metrics are effective for determining \textit{audio offsets} and identifying speakers in multi-person videos. 
Limitations of LSE-C and LSE-D include (1) highly limited correlation with human evaluation \citep{10647543}, (2) unreasonable values, \textit{e.g.,} ground truth was widely outperformed by generation results \citep{xu2024vasa}, (3) instability to factors such as the cropped mouth region, image quality and brightness \citep{yaman2024audio,yaman2024audio_2}.

\paragraph{User studies.} Given the above limitations of objective metrics, user studies remain a viable evaluation for TH generation. However, associated limitations include the tedious and time-consuming nature of such studies.\\ 

Current methods often use a few limited metrics
, to summarize the complex process of generating TH. Relying on single metrics can render aspects related to over or underperformance not interpretable. Such aspects can be of different nature, 
\textit{e.g.,} accurate mouth movements, realistic head motion, as well as overall appearance. 

Motivated by the above gap, we propose \textsc{THEval}, a framework developed to enhance existing metrics for evaluating TH videos. Rather than relying on a limited set of general metrics for assessing performance of TH-generation models, we advocate for a detailed breakdown of relevant factors, encompassing global head movement to nuanced expression. This decomposition facilitates a more comprehensive understanding, enabling targeted improvements in future models. 
\section{THEval Metrics}
\label{sec:Method}

\begin{figure*}[t!] 
    \centering
    \includegraphics[width=0.9\textwidth]{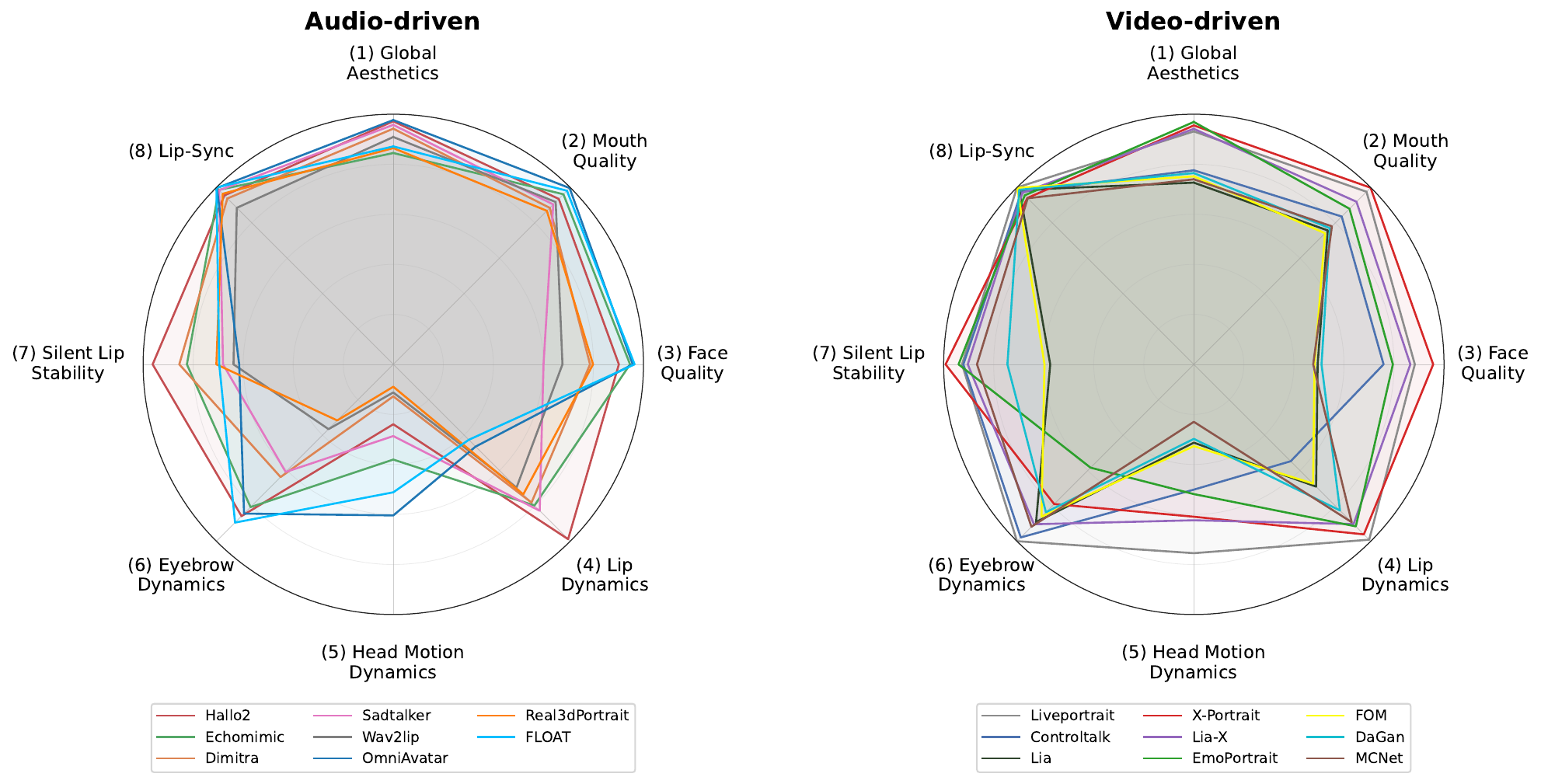}
    
    \caption{\textbf{Quantitative comparison of audio-driven (left) and video-driven (right) models on the THEval framework.} The radar charts visualize performance across our eight evaluation metrics, revealing distinct performance profiles. Video-driven models generally achieve more balanced, high-quality results, while audio-driven models exhibit greater variance, often excelling in dynamics but struggling with overall naturalness. Scores that are farther from the center indicate superior performance.}
    \label{fig:spider_res}
\end{figure*}

We introduce our proposed \textsc{THEval}, which combines 8 algorithmic and perceptual metrics. Drawing from recent video generation benchmarks \citep{huang2023vbench}, our framework entails 3 core dimensions representing (i) \textit{quality}, (ii) \textit{naturalness}, as well as (iii) \textit{synchronization}. A visual representation of the results is available in Figure \ref{fig:spider_res}. We proceed with the motivation and implementation of each metric.

\subsection{Quality}
This dimension reflects the perceptual appeal of a generated TH, with emphasis on clarity, sharpness, and color fidelity. 

\paragraph{(1) Global Aesthetics.} We introduce a global aesthetics measure by adopting the Image Aesthetic Assessment (IAA) component of TOPIQ \citep{chen2024topiq}, which accounts for high-level attributes such as composition, lighting, and color harmony, with $S_{aes,j}$ denoting the aesthetic score and $N$ the total number of frames
\begin{equation} \textit{Global Aesthetics} = \frac{1}{N} \sum_{j=1}^{N} S_{aes,j}. \end{equation}

\paragraph{(2), (3) Mouth- and Face-Centric Quality.} We propose a region-specific assessment strategy. Specifically, we compute Image Quality Assessment (IQA) with TOPIQ for the full face and MUSIQ \citep{ke2021musiq} for the mouth region. To this end, we localize facial regions via landmarks and extract IQA scores separately, with $Q$ denoting the quality score for the face and mouth

\begin{equation} \textit{Face / Mouth Quality} = \frac{1}{N} \sum_{j=1}^{N} Q_{face / mouth,j}. \end{equation}

Separating the mouth from the face allows us to explicitly quantify the distinct difficulty of synthesizing realistic mouth motion in generative video models.


\subsection{Naturalness}

Naturalness indicates the realism of facial behavior in TH videos. Prior work \citep{hauser2024} shows that subtle facial asymmetries such as in eyebrow, mouth, or head motion enhance perceived believability, appeal, and naturalness. Motivated by these findings, we include metrics that quantify lip, eyebrow, and head movement dynamics to assess whether TH videos exhibit realistic and engaging facial behavior.

\paragraph{(4)} The \textbf{Lip Dynamics}\label{Method:MMD} metric quantifies the variability of lip movements. For each frame $j$, we extract $K=40$ lip landmarks and compute a feature vector $s_j$ consisting of $M$ pairwise Euclidean distances $d_{j,m}$ that capture the lip shape, we use the combination of all the lip landmark. The metric is the average standard deviation $\sigma_m$ of these distances across all $N$ frames:
\begin{equation}
\textit{Lip Dynamics} = \frac{1}{M} \sum_{m=1}^{M} \sigma_m
\end{equation}
where $\sigma_m = \sqrt{\frac{1}{N-1} \sum_{j=1}^{N} (d_{j,m} - \bar{d}_m)^2}$ is the standard deviation of the $m$-th distance component and $\bar{d}_m$ is its mean.

\paragraph{(5)} The \textbf{Head Motion Dynamics} metric quantifies head movements in a video. We first estimate the head's orientation, specifically, its pitch, yaw, and roll, as well as the position of the head within the frame. For each video segment, the metric follows changes in head pose angles and translations over time. It is defined as follows, where $\overline{\sigma_{\text{angle}}}$ is the mean standard deviation of pitch, yaw, and roll angles, $\overline{V_{\Delta \text{angle}}}$ is the mean variance of their first-order temporal differences, and $\overline{V_{\text{trans}}}$ is the mean variance of face center translations
\begin{equation}
\textit{Head Motion Dyn.} = 
\sqrt{\bigl(\overline{\sigma_{\text{angle}}} \cdot \overline{V_{\Delta \text{angle}}}\bigr) + \overline{V_{\text{trans}}}}.
\end{equation}

\paragraph{(6)} The \textbf{Eyebrow Dynamics} metric captures the variability of eyebrow movements, which represent a wide range of emotional expressions \cite{wierzbicka2000semantics}. For each frame $j$, a representative vertical distance $d_{eb,j}$ between the eyebrows and eyes is calculated. This is normalized by the inter-ocular distance $d_{io,j}$ (distance between eye centers) to account for face scale:
\begin{equation}
    d'_{eb,j} = \frac{d_{eb,j}}{d_{io,j}}
\end{equation}
The Eyebrow Dynamics score is the standard deviation of this normalized relative distance over all $N$ frames:
\begin{equation} 
\textit{Eyebrow Dynamics} = \sqrt{\frac{1}{N-1} \sum_{j=1}^{N} (d'_{eb,j} - \overline{d'_{eb}})^2} 
\end{equation}
where $\overline{d'_{eb}}$ is the mean of $d'_{eb,j}$ over all frames.

\begin{figure*}[htbp]
    \centering
    \includegraphics[width=\textwidth]{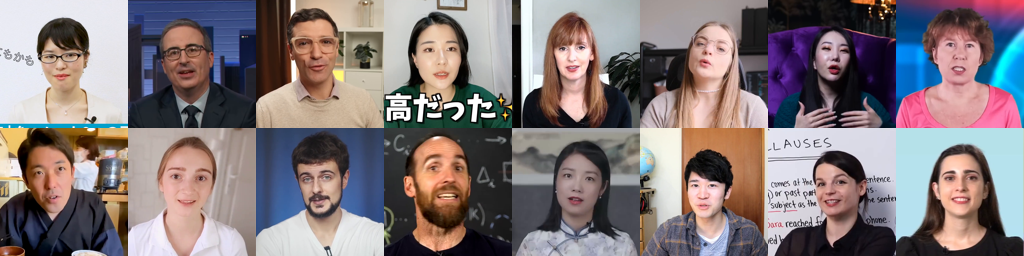}
    \caption{\textbf{Visual examples from our new THEval dataset.} Our benchmark is curated for diversity, featuring a wide range of subjects, head poses, and expressions from multiple linguistic backgrounds (including Spanish, Italian, English, French, Japanese, and Chinese). This dataset is specifically designed to test the generalization capabilities of talking head generation models on truly unseen data.}
    \label{fig:dataset_samples_alt}
\end{figure*}

\subsection{Synchronization}

Synchronization assesses the extent to which a TH's lip movements aligns with the accompanying audio-speech. Notably, Chae-Yeon \textit{et al.} \citep{chae2025perceptually} indicated that viewers are sensitive to related misalignment and tend to favor scenarios, where the intensity of lip and jaw movements corresponds to the audio volume. Building on these findings, we incorporate metrics that evaluate lip stability during silent intervals and the alignment of mouth openness with speech intensity to determine whether TH videos demonstrate realistic and expressive lip synchronization.

\paragraph{(7)} The \textbf{Silent Lip Stability} relates to mouth movement during silent periods, using a robust estimator. First, silent segments $S_{\text{silent}}$ (duration $\geq 300$\,ms) are identified using a Voice Activity Detection model (VAD). For each frame $j \in S_{\text{silent}}$, facial landmarks are detected, and we compute the per-frame average mouth opening $d_{lip,j}$ by averaging $P$ pre-defined vertical lip distances, normalized by the inter-ocular distance $d_{io,j}$:
\begin{equation}
    d_{lip,j} = \frac{1}{P} \sum_{p=1}^{P} \frac{|y_{\text{upper},p,j} - y_{\text{lower},p,j}|}{d_{io,j}}
\end{equation}
Stability is quantified using the Median Absolute Deviation (MAD):
\begin{equation}
    \textit{Silent Lip Stability} = \text{median}\big( \big| d_{lip,j} - \tilde{d}_{lip} \big| \big)
\end{equation}
where $\tilde{d}_{lip}$ is the median of $d_{lip,j}$ over all $j \in S_{\text{silent}}$.

\paragraph{(8)} The \textbf{Lip-Sync} metric evaluates the degree of alignment between mouth movements and the spoken audio. First, we use the same VAD model as \textit{(7) Silent Lip Stability} to identify frames containing speech, $S$. For each frame $t \in S$, we compute the mouth openness $O_t$ and the corresponding audio volume $V_t$ from the Root-Mean-Square (RMS) energy. Specifically, mouth openness $O_t$ is the vertical distance between the mean position of the upper and lower lip landmarks, normalized by the inter-ocular distance $d_{io,t}$:
\begin{equation}
    O_t = \frac{\big|\bar{y}_{\text{upper},t} - \bar{y}_{\text{lower},t}\big|}{d_{io,t}}
\end{equation}
Both the mouth openness ($O_S$) and audio volume ($V_S$) signals are independently min-max normalized to a [0, 1] range:
\begin{equation}
    O^*_t = \textstyle\frac{O_t - \min(O_S)}{\max(O_S) - \min(O_S)}, \qquad \\
    V^*_t = \textstyle\frac{V_t - \min(V_S)}{\max(V_S) - \min(V_S)}
\end{equation}
where $O_S$ and $V_S$ are the sets of mouth openness and volume values for all frames in $S$. The final lip-sync metric $L_{sync}$ is the mean absolute difference between these two normalized signals:
\begin{equation}
    L_{sync} = \frac{1}{|S|} \sum_{t \in S} \left| O^*_t - V^*_t \right|
\end{equation}

\begin{table*}[ht]
\centering
{ 
\begin{adjustbox}{width=\textwidth}
\begin{tabular}{l l c c c c c}
\toprule
& \textbf{Model} & \multicolumn{3}{c}{\textbf{Quality}} & \multicolumn{2}{c}{\textbf{Synchronization}} \\
\cmidrule(lr){3-5} \cmidrule(lr){6-7}
& & \textbf{(1) Global Aesthetics $\uparrow$} & \textbf{(2) Mouth Quality $\uparrow$} & \textbf{(3)Face Quality $\uparrow$} & \textbf{(7) Silent Lip Stability $\uparrow$} & \textbf{(8) Lip-Sync $\uparrow$} \\
\midrule
\multirow{7}{*}{\rotatebox{90}{\shortstack{\textbf{Audio}\\\textbf{Driven}}}}
& Hallo2~\citep{cui2024hallo2} & \textbf{0.9619} & 0.9254 & 0.9017 & \textbf{0.9620} & 0.9502 \\
& OmniAvatar~\citep{gan2025omniavatar} & \textbf{0.9767} & \textbf{0.9919} & \textbf{0.9521} & 0.6160 & \textbf{0.9972} \\
& Echomimic~\citep{chen2024echomimic} & 0.8499 & \textbf{0.9617} & \textbf{0.9514} & \textbf{0.8251} & \textbf{0.9964} \\
& FLOAT~\citep{ki2024float} & 0.8713 & \textbf{0.9868} & \textbf{0.9645} & 0.6958 & \textbf{0.9992} \\
& Sadtalker~\citep{zhang2023sadtalker} & \textbf{0.9576} & 0.9142 & 0.6005 & 0.6806 & 0.9794 \\
& Dimitra~\citep{chopin2025Dimitra} & 0.9523 & 0.8798 & 0.7914 & \textbf{0.8555} & 0.9430 \\
& Real3dPortrait~\citep{ye2024real3d} & 0.8597 & 0.8732 & 0.7934 & 0.7072 & 0.9695 \\
& Wav2lip~\citep{wav2lips} & 0.9090 & 0.9180 & 0.6762 & 0.6388 & 0.8849 \\
\midrule
\multirow{10}{*}{\rotatebox{90}{\shortstack{\textbf{Video}\\\textbf{Driven}}}}
& LIA-X~\citep{wang2025lia} & \textbf{0.9466} & \textbf{0.9195} & \textbf{0.8705} & 0.9087 & 0.9644 \\
& Liveportrait~\citep{guo2024liveportrait} & 0.9464 & \textbf{0.9760} & \textbf{0.8784} & \textbf{0.9316} & \textbf{0.9980} \\
& X-Portrait~\citep{xie2024x} & \textbf{0.9502} & \textbf{0.9990} & \textbf{0.9568} & \textbf{0.9924} & 0.9407 \\
& EmoPortrait~\citep{drobyshev2024emoportraits} & \textbf{0.9542} & 0.8799 & 0.7957 & \textbf{0.9354} & 0.9608 \\
& Controltalk~\citep{zhao2024controllable} & 0.7759 & 0.8360 & 0.7584 & 0.9163 & 0.9897 \\
& MCNet~\citep{mcnet} & 0.7499 & 0.7655 & 0.4771 & 0.8669 & 0.9541 \\
& DaGan~\citep{dagan} & 0.7547 & 0.7646 & 0.5105 & 0.7452 & 0.9719 \\
& LIA~\citep{wang2022latent} & 0.7265 & 0.7622 & 0.4899 & 0.5741 & \textbf{0.9913} \\
& FOM~\citep{fom} & 0.7516 & 0.7566 & 0.4875 & 0.5970 & \textbf{0.9929} \\
\bottomrule
\end{tabular}
\end{adjustbox}
\tiny
\begin{adjustbox}{width=\textwidth}
\begin{tabular}{l l c c c c}
\toprule[0.06em]
& \textbf{Model} & \multicolumn{3}{c}{\textbf{Naturalness}} & \textbf{Final Score $\uparrow$} \\
\cmidrule(lr){3-5} \cmidrule(lr){6-6}
& & \textbf{(4) Lip Dynamics $\uparrow$} & \textbf{(5) Head Motion $\uparrow$} & \textbf{(6) Eyebrow Dynamics $\uparrow$} & \\
\midrule[0.06em]
\multirow{7}{*}{\rotatebox{90}{\shortstack{\textbf{Audio}\\\textbf{Driven}}}}
& Hallo2~\citep{cui2024hallo2} & \textbf{0.9883} & 0.2395 & \textbf{0.8530} & \textbf{0.8477} \\
& OmniAvatar~\citep{gan2025omniavatar} & 0.4650 & \textbf{0.6039} & \textbf{0.8488} & \textbf{0.8064} \\
& Echomimic~\citep{chen2024echomimic} & \textbf{0.7930} & \textbf{0.3806} & 0.8071 & \textbf{0.8207} \\
& FLOAT~\citep{ki2024float} & 0.4266 & \textbf{0.5115} & \textbf{0.8945} & 0.7938 \\
& Sadtalker~\citep{zhang2023sadtalker} & \textbf{0.8276} & 0.2867 & 0.6084 & 0.7319 \\
& Dimitra~\citep{chopin2025Dimitra} & 0.7863 & 0.1279 & 0.6372 & 0.7467 \\
& Real3dPortrait~\citep{ye2024real3d} & 0.7348 & 0.0895 & 0.3170 & 0.6680 \\
& Wav2lip~\citep{wav2lips} & 0.6966 & 0.1124 & 0.3662 & 0.6502 \\
\midrule[0.06em]
\multirow{10}{*}{\rotatebox{90}{\shortstack{\textbf{Video}\\\textbf{Driven}}}}
& LIA-X~\citep{wang2025lia} & 0.9030 & \textbf{0.6233} & 0.9090 & \textbf{0.8806} \\
& Liveportrait~\citep{guo2024liveportrait} & \textbf{0.9913} & \textbf{0.7548} & \textbf{0.9997} & \textbf{0.9345} \\
& X-Portrait~\citep{xie2024x} & \textbf{0.9611} & \textbf{0.6091} & 0.7897 & \textbf{0.8999} \\
& EmoPortrait~\citep{drobyshev2024emoportraits} & \textbf{0.9159} & 0.5136 & 0.5840 & 0.8174 \\
& Controltalk~\citep{zhao2024controllable} & 0.5476 & 0.5058 & \textbf{0.9785} & 0.7885 \\
& MCNet~\citep{mcnet} & 0.8925 & 0.2297 & \textbf{0.9132} & 0.7311 \\
& DaGan~\citep{dagan} & 0.8262 & 0.3029 & 0.8362 & 0.7140 \\
& LIA~\citep{wang2022latent} & 0.6912 & 0.3080 & 0.8920 & 0.6794 \\
& FOM~\citep{fom} & 0.6743 & 0.3269 & 0.8613 & 0.6810 \\
\bottomrule[0.06em]
\end{tabular}
\end{adjustbox}
\caption{\textbf{Evaluation results of 17 TH Generation Models on the THEval Dataset.} We compare state-of-the-art audio-driven and video-driven models across our proposed categories of (i) \textit{quality}, (ii) \textit{synchronization}, and (iii) \textit{naturalness}. The results highlight the distinct performance profiles of the two approaches. For each metric, higher values indicate better performance. The scores in bold represent the three best scores per dimension.}
} 
\label{tab:results}
\end{table*}

\paragraph{Final Score} To provide a single, interpretable measure for evaluating TH-videos, we compute a \textit{final composite score} that aggregates performance across the three key dimensions (i) \textit{quality}, (ii) \textit{Naturalness}, and (iii) \textit{Synchronization}. This score is derived from the eight metrics that we introduced, which are first normalized relative to a ground-truth (GT) reference and then averaged into a single value. For normalization, a model's score on a specific metric is determined by its similarity to the GT value, as expressed by the following equation. 
\begin{equation}
    s = 1 - \frac{| \text{Model}_{Score} - \text{GT}_{Score} |}{\text{GT}_{Score}}
\end{equation}
where a score of $1$ indicates a perfect match with the ground truth, lower values reflect increasing deviation. We compute the final score using an unweighted average of the normalized metrics. This choice is justified by the strong, significant alignment with human preferences already achieved ($\rho$ = 0.870). Although weighting methods (e.g., linear regression, correlation scores) could maximize correlation on our specific test set, we prioritize the unweighted approach for transparency and robustness. It ensures clarity while maintaining the equal and complementary importance of the Quality, Naturalness, and Synchronization dimensions, avoiding arbitrary bias or overfitting to the specific human evaluation sample.

\section{Experiments}
\label{Exp}

We present a series of experiments designed to validate the effectiveness of our proposed evaluation framework, including the new evaluation dataset, the methods compared, and the correlation with human ratings.

\subsection{THEval Dataset}
\label{sec:dataset}

To thoroughly assess the generalization capabilities of contemporary TH models, we present the \textsc{THEval dataset}, a new benchmark designed to highlight the benefits and limitations of the models under evaluation. Our primary goal was to create an evaluation set with samples explicitly not seen during the training of the models we evaluate. The dataset was constructed by sourcing 5,011 video from a wide range of 31 public YouTube channels, ensuring a rich variety of content across multiple languages, including Spanish, Italian, English, French, Japanese, and Chinese. Each video have a single speaker, a clear and primarily frontal view of the face, and high-fidelity 1080p resolution. This resulted in a final dataset of over 18 hours of content, with an average video length of 13 seconds, more details about the dataset are available in SM~\ref{app:dataset_details}. Visual examples showcasing this diversity are presented in Figure~\ref{fig:dataset_samples_alt}.

\subsection{Setup}
\label{subsec:setup}

\paragraph{Compared Methods.}
We evaluate the following 17 state-of-the-art TH generation models for video-driven approaches,
\textbf{Controltalk}~\citep{zhao2024controllable},
\textbf{Liveportrait}~\citep{guo2024liveportrait},
\textbf{LIA}~\citep{wang2022latent}, \textbf{X-Portrait}~\citep{xie2024x}, \textbf{LIA-X}~\citep{wang2025lia}, \textbf{EmoPortrait}~\citep{drobyshev2024emoportraits}, \textbf{MCNet}~\citep{mcnet}, \textbf{DaGan}~\citep{dagan}, \textbf{FOM}~\citep{fom} and audio-driven approaches
\textbf{Hallo2}~\citep{cui2024hallo2},
\textbf{EchoMimic}~\citep{chen2024echomimic},
\textbf{Wav2Lip}~\citep{wav2lips}, \textbf{SadTalker}~\citep{zhang2023sadtalker}, \textbf{Dimitra}~\citep{chopin2025Dimitra}, \textbf{OmniAvatar}~\citep{gan2025omniavatar}, \textbf{Real3dPortrait}~\citep{ye2024real3d} and \textbf{FLOAT}~\citep{ki2024float}.

Details associated to the above methods are provided in SM \ref{methods_desc}. Each method is executed with its default hyperparameters, and model weights provided by the authors or official repositories. For a fair comparison, we provide the same audio and reference frames for each method, in order to generate videos. We do not evaluate 3D Gaussian Splatting or Neural Radiance Fields methods, as they require multi-view inputs and are limited to a fixed set of pretrained identities, making them unsuitable for our single-view, multi-identity setting.

\paragraph{Implementation Details.}
\label{implementation}
We employ MediaPipe Face Mesh~\citep{48292} for extracting facial landmarks, including eyes, lips, and eyebrows, across all frames. Head pose metrics are computed using FaceXFormer~\citep{narayan2024facexformer}. 
Finally, to detect speech segments in the audio we use Silero VAD \citep{SileroVAD}.

\paragraph{Ground Truth as Reference.}
We include comparisons with ground-truth GT videos as reference, in order to compute absolute differences with the generation methods. This allows us to quantitatively evaluate the resemblance of each method to real facial behavior. For instance, in metrics such as (5) Head Motion Dynamics, a low score indicates that the generated head movements are too subtle as opposed to ground truth, whereas a high score suggests exaggerated head motions.

\subsection{THEval Correlation with Human Rating}
\label{subsec:metric_val}

Towards validating our metrics, we conduct a user study in Hugging Face Space. Participants are asked to evaluate paired real and generated videos based on the same audio. Pairing included all combinations, \textit{i.e.,} real and generated videos, as well as generated videos pertained to all combinations of the seventeen state-of-the-art methods. Participants were instructed by \textit{"Please watch both videos and select which one looks more realistic"}. The instruction aims at minimizing cognitive load and reducing rating errors, ensuring participants could make consistent and intuitive judgments. In total, we acquired 3,519 ratings, distributed equally among the seventeen models. We then compute the Spearman correlation coefficient between normalized metric scores and human preference scores. We observe strong alignment between our THEval metrics and human opinion scores, as reported in Figure~\ref{fig:corr} and Table~\ref{tab:metric_correlation_single}. The \textit{final THEval score} is highly correlated with human ratings, with a strong correlation of $\rho$ = 0.870. In addition, the individual metrics exhibit high correlations with human preferences. We note that (4) Lip Dynamics and (8) Lip-Sync are of lower correlation, which is expected as human perception of realism integrates multiple cues reflected by the 8 metrics. By combining complementary \textit{expert} metrics, our framework achieves a strong aggregated alignment with human preference, advocating for the composite design of THEval. 

\begin{table}[t]
    \centering
    \small
    \begin{adjustbox}{width=\columnwidth}
    \begin{tabular}{@{}lccc@{}}
        \toprule
        \textbf{Metric} & \textbf{Correlation ($\rho$)} & \textbf{p-value} & \textbf{95\% CI} \\
        \midrule
        LSE-C & -0.164 & 0.530 & [-0.613, 0.388] \\
        LSE-D & -0.269 & 0.297 & [-0.675, 0.282] \\
        FVD   & 0.289  & 0.260 & [-0.321, 0.782] \\
        FID   & 0.210  & 0.416 & [-0.344, 0.710] \\
        LMD-F & 0.231  & 0.389 & [-0.392, 0.775] \\
        LMD-L & 0.227  & 0.399 & [-0.389, 0.759] \\
        \midrule
        (1) Global Aesthetics & 0.544 & 0.020 & [0.129, 0.795] \\
        (2) Mouth Visual Quality & 0.765 & $<0.001$ & [0.498, 0.917] \\
        (3) Face Quality & 0.699 & 0.001 & [0.430, 0.875] \\
        (4) Lip Dynamics & 0.414 & 0.088 & [-0.155, 0.769] \\
        (5) Head Motion Dynamics & 0.763 & $<0.001$ & [0.418, 0.942] \\
        (6) Eyebrow Dynamics & 0.527 & 0.025 & [0.060, 0.856] \\
        (7) Silent Lip Stability & 0.484 & 0.042 & [0.033, 0.808] \\
        (8) Lip-Sync & 0.404 & 0.097 & [-0.143, 0.775] \\
        \midrule
        Quality & 0.713 & 0.001 & [0.424, 0.895] \\
        Naturalness & 0.702 & 0.001 & [0.217, 0.862] \\
        Synchronization & 0.603 & 0.008 & [0.323, 0.919] \\
        \midrule
        Final Score & 0.870 & $<0.0001$ & [0.648, 0.967] \\
        \bottomrule
    \end{tabular}
    \end{adjustbox}
    \caption{\textbf{Correlation between metrics and human ratings.} 
    We report Spearman's rank correlation coefficient ($\rho$), p-values, and 95\% Confidence Intervals (CI) between each metric and human preferences. 
    The 95\% CIs are obtained via bootstrapping with $n = 10{,}000$ resamples. 
    The results clearly show that our proposed metrics and final scores have a strong, significant alignment with human ratings.}
    \label{tab:metric_correlation_single}
    \vspace{-0.4cm}
\end{table}

\begin{figure}
    \centering
    \includegraphics[scale=0.21]{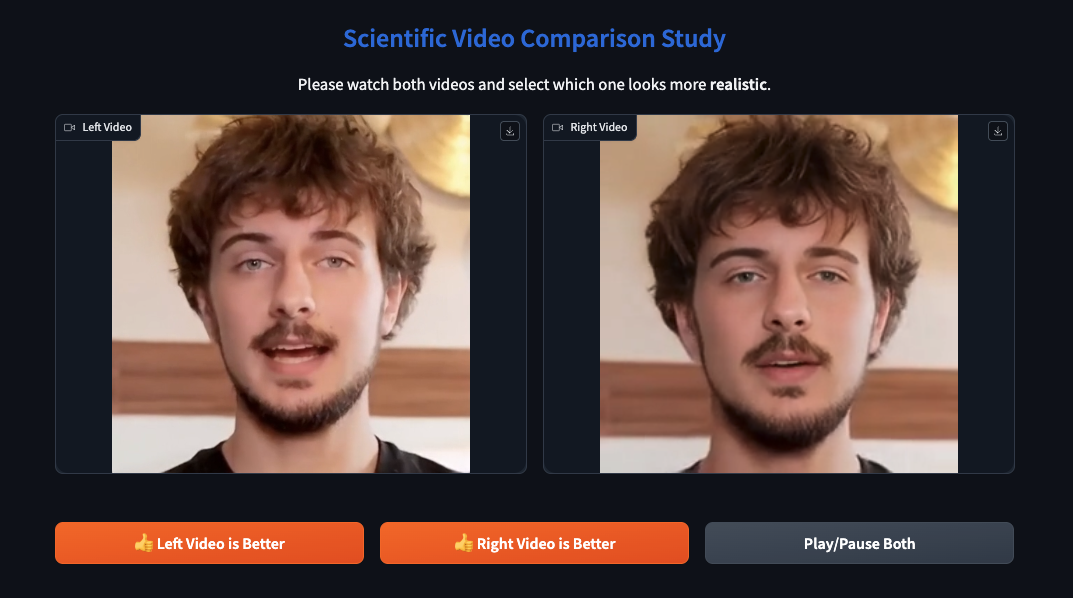}
    \caption{\textbf{Screenshot of our user study interface.} The interface was designed to be intuitive and easy to use for human raters. Both videos can be played simultaneously using the \textit{Play/Pause Both} button, and participants indicate which video appears more realistic by selecting one of the two highlighted choice buttons.}
    \label{fig:website}
\end{figure}

In contrast, SyncNet, landmark-based metrics, FID, and FVD demonstrate minor to no alignment with human ratings. These results indicate that our approach provides a reliable alternative to current evaluation metrics.  We observe a negative correlation between the SyncNet-derived metrics (LSE-D, LSE-C) and human preference. This is because Wav2Lip, trained with SyncNet as a discriminator, achieves top algorithmic scores by optimizing the metric directly, yet it was among the least selected methods in the user study.

Further details on the user study including and algorithm details to ensure fair evaluation are in SM~\ref{sec:appendix_user_study}, the user study interface is presented in Figure~\ref{fig:website}.
\section{Discussion}

\subsection{Audio-Driven Methods}

We note that audio-driven approaches consistently struggle \textit{w.r.t.} facial expressiveness, as well as head pose movement, which is reflected in the videos. In addition, FLOAT and OmniAvatar tend to exaggerate mouth movements. In the case of OmniAvatar, this is due to the employed image-to-video model WanVideo~\citep{wan2025}, which generates exaggerated articulations that appear unnatural, reflected in the scores of metrics (4) Lip Dynamics and (7) Silent Lip Stability. Furthermore, we observe that in longer videos, OmniAvatar exhibits temporal drift, with the facial identity gradually diverging from the source, visual artifacts becoming increasingly visible, and a perceptible orange color emerge, all of which reduce overall video quality. This temporal instability prevents OmniAvatar from achieving the highest ranking in our benchmark, despite its otherwise strong performance. Visual examples are availabe in SM~\ref{app:omni_drift}.

Naturally, newer audio-driven methods perform well \textit{w.r.t.} overall video quality and visual appeal compared to earlier approaches, thanks to advances in backbone architectures and training data. However, amplified expressions pose a remaining challenge. 
By disentangling synchronization, expressiveness, and quality, THEval effectively highlights these nuanced behavior. 

\subsection{Video-Driven Methods}
Video-driven methods, in contrast to audio-driven ones, exhibit stronger expressivity, while maintaining reliable synchronization. This is due to the fact that motion priors from driving videos allow for generation of realistic head dynamics and subtle facial expressions. We observe that earlier video-driven models incorporate visible artifacts, particularly in case of large head movements. Such artifacts manifest as tearing, blur, or instability in the facial regions, leading to lower overall quality scores in our benchmark. This trade-off between expressivity and quality is effectively highlighted by our proposed metrics, which separate naturalness-related measures from face and mouth quality assessments. 

Overall, the best-performing video-driven models showcase a strong balance between expressivity, synchronization, and visual fidelity, obtaining a higher \textit{Final Score} in THEval. Our results indicate that the \textit{Final Score} reflects on these trade-offs, ranking methods based on human perception.

\section{Conclusions}
\label{sec:conclusion}
We introduced a comprehensive benchmark, referred to as \textsc{THEval}, streamlined to evaluate generated talking head (TH) videos. The eight metrics in THEval cover the three key dimensions (i) quality, (ii) naturalness, and (iii) synchronization. Our benchmark enables both, fine-grained and efficient assessment of TH videos. Experiments on a new evaluation dataset show that \textsc{THEval} metrics correlate strongly with human preference, unlike currently used metrics, which often fail to reflect perceptual ratings. We further observe that state-of-the-art audio- and video-driven generative models still face challenges in producing realistic lip movements, natural expressiveness, and artifact-free rendering. Specifically, audio-driven methods have advanced \textit{w.r.t.} synchronization, however often lack natural head motion and may incorporate exaggerated expressions. At the same time recent video-driven counterparts generate more expressive and realistic videos. Our THEval metrics capture this illustratively, and by aggregating them into a \textit{Final Score}, THEval provides both, detailed diagnostics and a final measure that matches human preference, rendering it a necessary benchmark for TH generation. We aim at fostering the development of new generation methods. Future work will extend the benchmark to more diverse scenarios, such as multiple humans and side views. 
\section{Acknowledgement}
\label{sec:acknowledgement}
This project was provided with computing HPC and storage resources by GENCI at IDRIS thanks to the grant 2024-AD011015982 on the supercomputer Jean Zay’s H100 partition.

{   
    \small
    \bibliographystyle{ieeenat_fullname}
    \bibliography{main}

@String(CVPR= {IEEE Conf. Comput. Vis. Pattern Recog.})

@String(ICCV= {Int. Conf. Comput. Vis.})

@String(ECCV= {Eur. Conf. Comput. Vis.})

@String(ICIP = {IEEE Int. Conf. Image Process.})

@String(ICLR = {Int. Conf. Learn. Represent.})

@String(IJCAI = {IJCAI})

@String(AAAI = {AAAI})

@String(CVPRW= {IEEE Conf. Comput. Vis. Pattern Recog. Worksh.})

@String(CVPR  = {CVPR})

@String(ICCV  = {ICCV})

@String(ECCV  = {ECCV})

@String(ICIP  = {ICIP})

@String(ICLR  = {ICLR})

@String(CVPRW= {CVPRW})

@inproceedings{xu2024ufogen,
  title={Ufogen: You forward once large scale text-to-image generation via diffusion gans},
  author={Xu, Yanwu and Zhao, Yang and Xiao, Zhisheng and Hou, Tingbo},
  booktitle={Proceedings of the IEEE/CVF Conference on Computer Vision and Pattern Recognition (CVPR)},
  year={2024}
}

@article{yang2024improving,
  title={Improving diffusion-based image synthesis with context prediction},
  author={Yang, Ling and Liu, Jingwei and Hong, Shenda and Zhang, Zhilong and Huang, Zhilin and Cai, Zheming and Zhang, Wentao and Cui, Bin},
  journal={Advances in Neural Information Processing Systems},
  volume={36},
  year={2024}
}

@inproceedings{shi2024instantbooth,
  title={Instantbooth: Personalized text-to-image generation without test-time finetuning},
  author={Shi, Jing and Xiong, Wei and Lin, Zhe and Jung, Hyun Joon},
  booktitle={Proceedings of the IEEE/CVF Conference on Computer Vision and Pattern Recognition (CVPR)},
  year={2024}
}

@article{chen2024subject,
  title={Subject-driven text-to-image generation via apprenticeship learning},
  author={Chen, Wenhu and Hu, Hexiang and Li, Yandong and Ruiz, Nataniel and Jia, Xuhui and Chang, Ming-Wei and Cohen, William W},
  journal={Advances in Neural Information Processing Systems},
  volume={36},
  year={2024}
}

@inproceedings{wang2024recipe,
  title={A recipe for scaling up text-to-video generation with text-free videos},
  author={Wang, Xiang and Zhang, Shiwei and Yuan, Hangjie and Qing, Zhiwu and Gong, Biao and Zhang, Yingya and Shen, Yujun and Gao, Changxin and Sang, Nong},
  booktitle={Proceedings of the IEEE/CVF Conference on Computer Vision and Pattern Recognition (CVPR)},
  year={2024}
}

@inproceedings{qing2024hierarchical,
  title={Hierarchical spatio-temporal decoupling for text-to-video generation},
  author={Qing, Zhiwu and Zhang, Shiwei and Wang, Jiayu and Wang, Xiang and Wei, Yujie and Zhang, Yingya and Gao, Changxin and Sang, Nong},
  booktitle={Proceedings of the IEEE/CVF Conference on Computer Vision and Pattern Recognition (CVPR)},
  year={2024}
}

@inproceedings{liu2024evalcrafter,
  title={Evalcrafter: Benchmarking and evaluating large video generation models},
  author={Liu, Yaofang and Cun, Xiaodong and Liu, Xuebo and Wang, Xintao and Zhang, Yong and Chen, Haoxin and Liu, Yang and Zeng, Tieyong and Chan, Raymond and Shan, Ying},
  booktitle={Proceedings of the IEEE/CVF Conference on Computer Vision and Pattern Recognition (CVPR)},
  year={2024}
}

@inproceedings{huang2023vbench,
  title={Vbench: Comprehensive benchmark suite for video generative models},
  author={Huang, Ziqi and He, Yinan and Yu, Jiashuo and Zhang, Fan and Si, Chenyang and Jiang, Yuming and Zhang, Yuanhan and Wu, Tianxing and Jin, Qingyang and Chanpaisit, Nattapol and others},
  booktitle={Proceedings of the IEEE/CVF Conference on Computer Vision and Pattern Recognition (CVPR)},
  year={2024}
}

@InProceedings{Chung16a,
  author       = "Chung, J.~S. and Zisserman, A.",
  title        = "Out of time: automated lip sync in the wild",
  booktitle    = "Proceedings of the Asian Conference on Computer Vision Workshops (ACCVW)",
  year         = "2016",
}

@inproceedings{
xu2024vasa,
title={{VASA}-1: Lifelike Audio-Driven Talking Faces Generated in Real Time},
author={Sicheng Xu and Guojun Chen and Yu-Xiao Guo and Jiaolong Yang and Chong Li and Zhenyu Zang and Yizhong Zhang and Xin Tong and Baining Guo},
booktitle={Proceedings of the Conference on Neural Information Processing Systems (NeurIPS)},
year={2024},
}

@article{chen2024echomimic,
  title={Echomimic: Lifelike audio-driven portrait animations through editable landmark conditions},
  author={Chen, Zhiyuan and Cao, Jiajiong and Chen, Zhiquan and Li, Yuming and Ma, Chenguang},
  journal={arXiv preprint arXiv:2407.08136},
  year={2024}
}

@INPROCEEDINGS{10647543,

  author={Zhang, Weixia and Zhu, Chengguang and Gao, Jingnan and Yan, Yichao and Zhai, Guangtao and Yang, Xiaokang},

  booktitle={Proceedings of the IEEE International Conference on Image Processing (ICIP)}, 

  title={A Comparative Study of Perceptual Quality Metrics For Audio-Driven Talking Head Videos}, 

  year={2024},
  }

@inproceedings{yaman2024audio,
  title={Audio-driven Talking Face Generation with Stabilized Synchronization Loss},
  author={Yaman, Dogucan and Eyiokur, Fevziye Irem and B{\"a}rmann, Leonard and Ekenel, Haz{\i}m Kemal and Waibel, Alexander},
  booktitle={Proceedings of the European Conference on Computer Vision (ECCV)},
  year={2024}
}

@inproceedings{wav2lips,
author = {Prajwal, K R and Mukhopadhyay, Rudrabha and Namboodiri, Vinay P. and Jawahar, C.V.},
title = {A Lip Sync Expert Is All You Need for Speech to Lip Generation In the Wild},
year = {2020},
booktitle = {Proceedings of the  ACM International Conference on Multimedia (ACM-MM)},
}

@article{ma2023dreamtalk,
title={Dreamtalk: When expressive talking head generation meets diffusion probabilistic models},
author={Ma, Yifeng and Zhang, Shiwei and Wang, Jiayu and Wang, Xiang and Zhang, Yingya and Deng, Zhidong},
journal={arXiv preprint arXiv:2312.09767},
year={2023}
}

@inproceedings{zhang2023sadtalker,
  title={SadTalker: Learning Realistic 3D Motion Coefficients for Stylized Audio-Driven Single Image Talking Face Animation},
  author={Zhang, Wenxuan and Cun, Xiaodong and Wang, Xuan and Zhang, Yong and Shen, Xi and Guo, Yu and Shan, Ying and Wang, Fei},
  booktitle={Proceedings of the IEEE/CVF Conference on Computer Vision and Pattern Recognition (CVPR)},
  year={2023}
}

@article{chopin2025Dimitra,
      title={Dimitra: Audio-driven Diffusion model for Expressive Talking Head Generation}, 
      author={Baptiste Chopin and Tashvik Dhamija and Pranav Balaji and Yaohui Wang and Antitza Dantcheva},
      year={2025},
    journal={arXiv preprint arXiv:2502.17198},
}

@article{narayan2024facexformer,
  title={FaceXFormer: A Unified Transformer for Facial Analysis},
  author={Narayan, Kartik and VS, Vibashan and Chellappa, Rama and Patel, Vishal M},
  journal={arXiv preprint arXiv:2403.12960},
  year={2024}
}

@inproceedings{48292,title	= {MediaPipe: A Framework for Perceiving and Processing Reality},author	= {Camillo Lugaresi and Jiuqiang Tang and Hadon Nash and Chris McClanahan and Esha Uboweja and Michael Hays and Fan Zhang and Chuo-Ling Chang and Ming Yong and Juhyun Lee and Wan-Teh Chang and Wei Hua and Manfred Georg and Matthias Grundmann},year	= {2019},booktitle	= {Proceedings of the IEEE/CVF Conference on Computer Vision and Pattern Recognition Workshops (CVPRW)}}

@inproceedings{wang2024microcinema,
  title={Microcinema: A divide-and-conquer approach for text-to-video generation},
  author={Wang, Yanhui and Bao, Jianmin and Weng, Wenming and Feng, Ruoyu and Yin, Dacheng and Yang, Tao and Zhang, Jingxu and Dai, Qi and Zhao, Zhiyuan and Wang, Chunyu and others},
  booktitle={Proceedings of the IEEE/CVF Conference on Computer Vision and Pattern Recognition (CVPR)},
  year={2024}
}

@article{xue2024raphael,
  title={Raphael: Text-to-image generation via large mixture of diffusion paths},
  author={Xue, Zeyue and Song, Guanglu and Guo, Qiushan and Liu, Boxiao and Zong, Zhuofan and Liu, Yu and Luo, Ping},
  journal={Advances in Neural Information Processing Systems (NeurIPS)},
  volume={36},
  year={2024}
}

@InProceedings{Ni_2023_CVPR,
    author    = {Ni, Haomiao and Shi, Changhao and Li, Kai and Huang, Sharon X. and Min, Martin Renqiang},
    title     = {Conditional Image-to-Video Generation With Latent Flow Diffusion Models},
    booktitle = {Proceedings of the IEEE/CVF Conference on Computer Vision and Pattern Recognition (CVPR)},
    year      = {2023}}

@inproceedings{wang2022latent,
  title={Latent Image Animator: Learning to Animate Images via Latent Space Navigation},
  author={Wang, Yaohui and Yang, Di and Bremond, Francois and Dantcheva, Antitza},
  booktitle={Proceedings of the International Conference on Learning Representations (ICLR)},
  year={2022}
}

@inproceedings{Zhou2021Pose,
  title = {Pose-Controllable Talking Face Generation by Implicitly Modularized Audio-Visual Representation},
  author = {Zhou, Hang and Sun, Yasheng and Wu, Wayne and Loy, Chen Change and Wang, Xiaogang and Liu, Ziwei},
  booktitle = {Proceedings of the IEEE Conference on Computer Vision and Pattern Recognition (CVPR)},
  year = {2021}
}

@InProceedings{Wang_2023_CVPR,
    author    = {Wang, Duomin and Deng, Yu and Yin, Zixin and Shum, Heung-Yeung and Wang, Baoyuan},
    title     = {Progressive Disentangled Representation Learning for Fine-Grained Controllable Talking Head Synthesis},
    booktitle = {Proceedings of the IEEE/CVF Conference on Computer Vision and Pattern Recognition (CVPR)},
    year      = {2023},
}

@misc{gururani2022SPACE,
  title={{SPACE: Speech-driven Portrait Animation with Controllable Expression}},
  author={Siddharth Gururani and Arun Mallya and Ting-Chun Wang and Rafael Valle and Ming-Yu Liu},
  journal={Proceedings of the IEEE/CVF International Conference on Computer Vision (ICCV)},
  year={2023}
}

@InProceedings{wang2021audio2head,
author = {Suzhen, Wang and Lincheng, Li and Yu, Ding and Changjie, Fan and Xin, Yu},
title = {Audio2Head: Audio-driven One-shot Talking-head Generation with Natural Head Motion},
booktitle = {Proceedings of the  International Joint Conference on Artificial Intelligence (IJCAI)},
year = {2021},
}

@INPROCEEDINGS {oneshottalking,
author = {Z. Zhang and L. Li and Y. Ding and C. Fan},
booktitle = {Proceedings of the IEEE/CVF Conference on Computer Vision and Pattern Recognition (CVPR)},
title = {Flow-guided One-shot Talking Face Generation with a High-resolution Audio-visual Dataset},
year = {2021},
}

@article{thies2020nvp,
  author = {Thies, Justus and Elgharib, Mohamed and Tewari, Ayush and Theobalt, Christian and Nie{\ss}ner, Matthias},
  title = {Neural Voice Puppetry: Audio-driven Facial Reenactment},
  journal={Proceedings of the European Conference on Computer Vision (ECCV)},
  year={2020}
}

@INPROCEEDINGS {taklingheadgen,
author = {Z. Yu and Z. Yin and D. Zhou and D. Wang and F. Wong and B. Wang},
booktitle = {Proceedings of the IEEE/CVF International Conference on Computer Vision (ICCV)},
title = {Talking Head Generation with Probabilistic Audio-to-Visual Diffusion Priors},
year = {2023},
}

@inproceedings{yu2023talking,
  title={Talking head generation with probabilistic audio-to-visual diffusion priors},
  author={Yu, Zhentao and Yin, Zixin and Zhou, Deyu and Wang, Duomin and Wong, Finn and Wang, Baoyuan},
  booktitle={Proceedings of the IEEE/CVF International Conference on Computer Vision (ICCV)},
  year={2023}
}

@inproceedings{shen2023difftalk,
   author={Shen, Shuai and Zhao, Wenliang and Meng, Zibin and Li, Wanhua and Zhu, Zheng and Zhou, Jie and Lu, Jiwen},
   title={DiffTalk: Crafting Diffusion Models for Generalized Audio-Driven Portraits Animation},
   booktitle={Proceedings of the IEEE/CVF Conference on Computer Vision and Pattern Recognition (CVPR)},
   year={2023}
}

@InProceedings{Chong_2020_CVPR,
author = {Chong, Min Jin and Forsyth, David},
title = {Effectively Unbiased FID and Inception Score and Where to Find Them},
booktitle = {Proceedings of the IEEE/CVF Conference on Computer Vision and Pattern Recognition (CVPR)},
year = {2020}
}

@misc{1998RecIB,
  title={Rec. ITU-R BT.1359-1 1 RECOMMENDATION ITU-R BT.1359-1 RELATIVE TIMING OF SOUND AND VISION FOR BROADCASTING},
  author={},
  year={1998},
}

@misc{ma2023styletalk,
  title={Styletalk: One-shot talking head generation with controllable speaking styles},
  author={Ma, Yifeng and Wang, Suzhen and Hu, Zhipeng and Fan, Changjie and Lv, Tangjie and Ding, Yu and Deng, Zhidong and Yu, Xin},
  booktitle={Proceedings of the AAAI Conference on Artificial Intelligence (AAAI)},
  volume={37},
  year={2023}
}

@inproceedings{yaman2024audio_2,
  title={Audio-Visual Speech Representation Expert for Enhanced Talking Face Video Generation and Evaluation},
  author={Yaman, Dogucan and Eyiokur, Fevziye Irem and B{\"a}rmann, Leonard and Akti, Seymanur and Ekenel, Haz{\i}m Kemal and Waibel, Alexander},
  booktitle={Proceedings of the IEEE/CVF Conference on Computer Vision and Pattern Recognition Workshops (CVPRW)},
  year={2024}
}

@article{wang2022anyonenet,
  title={Anyonenet: Synchronized speech and talking head generation for arbitrary persons},
  author={Wang, Xinsheng and Xie, Qicong and Zhu, Jihua and Xie, Lei and Scharenborg, Odette},
  journal={IEEE Transactions on Multimedia},
  volume={25},
  year={2022},
}

@inproceedings{li2021write,
  title={Write-a-speaker: Text-based emotional and rhythmic talking-head generation},
  author={Li, Lincheng and Wang, Suzhen and Zhang, Zhimeng and Ding, Yu and Zheng, Yixing and Yu, Xin and Fan, Changjie},
  booktitle={Proceedings of the AAAI conference on artificial intelligence (AAAI)},
  volume={35},
  year={2021}
}

@article{wang2024lia,
  title={LIA: Latent Image Animator},
  author={Wang, Yaohui and Yang, Di and Bremond, Francois and Dantcheva, Antitza},
  journal={IEEE Transactions on Pattern Analysis and Machine Intelligence},
  year={2024},
  publisher={IEEE}
}

@article{chen2024topiq,
  author={Chen, Chaofeng and Mo, Jiadi and Hou, Jingwen and Wu, Haoning and Liao, Liang and Sun, Wenxiu and Yan, Qiong and Lin, Weisi},
  title={TOPIQ: A Top-Down Approach From Semantics to Distortions for Image Quality Assessment}, 
  journal={IEEE Transactions on Image Processing}, 
  year={2024},
  volume={33},
  pages={2404-2418},
  doi={10.1109/TIP.2024.3378466}
}

@inproceedings{ke2021musiq,
  title={MUSIQ: Multi-scale Image Quality Transformer},
  author={Ke, Junjie and Wang, Qifei and Wang, Yilin and Milanfar, Peyman and Yang, Feng},
  booktitle={Proceedings of the IEEE/CVF International Conference on Computer Vision},
  pages={5148--5157},
  year={2021}
}

@misc{SileroVAD,
  author = {{Silero Team}},
  title = {Silero VAD: pre-trained enterprise-grade Voice Activity Detector (VAD), Number Detector and Language Classifier},
  year = {2024},
  publisher = {GitHub},
  journal = {GitHub repository},
  howpublished = {\url{https://github.com/snakers4/silero-vad}},
  commit = {insert_some_commit_here},
  email = {hello@silero.ai}
}

@article{zhao2024controllable,
  title={Controllable Talking Face Generation by Implicit Facial Keypoints Editing},
  author={Zhao, Dong and Shi, Jiaying and Li, Wenjun and Wang, Shudong and Xu, Shenghui and Pan, Zhaoming},
  journal={arXiv preprint arXiv:2406.02880},
  year={2024}
}

@misc{cui2024hallo2,
	title={Hallo2: Long-Duration and High-Resolution Audio-driven Portrait Image Animation},
	author={Jiahao Cui and Hui Li and Yao Yao and Hao Zhu and Hanlin Shang and Kaihui Cheng and Hang Zhou and Siyu Zhu and Jingdong Wang},
	year={2024},
	eprint={2410.07718},
	archivePrefix={arXiv},
	primaryClass={cs.CV}
}

@article{guo2024liveportrait,
  title   = {LivePortrait: Efficient Portrait Animation with Stitching and Retargeting Control},
  author  = {Guo, Jianzhu and Zhang, Dingyun and Liu, Xiaoqiang and Zhong, Zhizhou and Zhang, Yuan and Wan, Pengfei and Zhang, Di},
  journal = {arXiv preprint arXiv:2407.03168},
  year    = {2024}
}

@article{wang2025lia,
  title={LIA-X: Interpretable Latent Portrait Animator},
  author={Wang, Yaohui and Yang, Di and Chen, Xinyuan and Bremond, Francois and Qiao, Yu and Dantcheva, Antitza},
  journal={arXiv preprint arXiv:2508.09959},
  year={2025}
}

@article{xie2024x,
  title={X-Portrait: Expressive Portrait Animation with Hierarchical Motion Attention},
  author={Xie, You and Xu, Hongyi and Song, Guoxian and Wang, Chao and Shi, Yichun and Luo, Linjie},
  journal={arXiv preprint arXiv:2403.15931},
  year={2024}
}

@misc{drobyshev2024emoportraits,
      title={EMOPortraits: Emotion-enhanced Multimodal One-shot Head Avatars}, 
      author={Nikita Drobyshev and Antoni Bigata Casademunt and Konstantinos Vougioukas and Zoe Landgraf and Stavros Petridis and Maja Pantic},
      year={2024},
      eprint={2404.19110},
      archivePrefix={arXiv},
      primaryClass={cs.CV}
}

@article{ye2024real3d,
  title={Real3D-Portrait: One-shot Realistic 3D Talking Portrait Synthesis},
  author={Ye, Zhenhui and Zhong, Tianyun and Ren, Yi and Yang, Jiaqi and Li, Weichuang and Huang, Jiawei and Jiang, Ziyue and He, Jinzheng and Huang, Rongjie and Liu, Jinglin and others},
  journal={arXiv preprint arXiv:2401.08503},
  year={2024}
}

@misc{gan2025omniavatar,
      title={OmniAvatar: Efficient Audio-Driven Avatar Video Generation with Adaptive Body Animation}, 
      author={Qijun Gan and Ruizi Yang and Jianke Zhu and Shaofei Xue and Steven Hoi},
      year={2025},
      eprint={2506.18866},
      archivePrefix={arXiv},
      primaryClass={cs.CV},
      url={https://arxiv.org/abs/2506.18866}, 
}

@inproceedings{mcnet,
            title={Implicit Identity Representation Conditioned Memory Compensation Network for Talking Head video Generation},
            author={Hong, Fa-Ting and Xu, Dan},
            booktitle={ICCV},
            year={2023}
}

@article{wan2025,
      title={Wan: Open and Advanced Large-Scale Video Generative Models}, 
      author={{Wan Team} and Ang Wang and Baole Ai and Bin Wen and Chaojie Mao and Chen-Wei Xie and Di Chen and Feiwu Yu and Haiming Zhao and Jianxiao Yang and Jianyuan Zeng and Jiayu Wang and Jingfeng Zhang and Jingren Zhou and Jinkai Wang and Jixuan Chen and Kai Zhu and Kang Zhao and Keyu Yan and Lianghua Huang and Mengyang Feng and Ningyi Zhang and Pandeng Li and Pingyu Wu and Ruihang Chu and Ruili Feng and Shiwei Zhang and Siyang Sun and Tao Fang and Tianxing Wang and Tianyi Gui and Tingyu Weng and Tong Shen and Wei Lin and Wei Wang and Wei Wang and Wenmeng Zhou and Wente Wang and Wenting Shen and Wenyuan Yu and Xianzhong Shi and Xiaoming Huang and Xin Xu and Yan Kou and Yangyu Lv and Yifei Li and Yijing Liu and Yiming Wang and Yingya Zhang and Yitong Huang and Yong Li and You Wu and Yu Liu and Yulin Pan and Yun Zheng and Yuntao Hong and Yupeng Shi and Yutong Feng and Zeyinzi Jiang and Zhen Han and Zhi-Fan Wu and Ziyu Liu},
      journal = {arXiv preprint arXiv:2503.20314},
      year={2025}
}

@inproceedings{hauser2024,
            title={The Effect of Dynamic Facial Asymmetries on the Perceived Believability, Appeal, and Naturalness of Animated Agents},
            author={Hauser, Klay Max and Mousas, Christos and Adamo, Nicoletta and Choi, Minsoo and Mayer, Richard and Zhao, Fangzheng},
            booktitle={ACM Symposium on Applied Perception},
            year={2024}
}

@inproceedings{chae2025perceptually,
      title={Perceptually Accurate 3D Talking Head Generation: New Definitions, Speech-Mesh Representation, and Evaluation Metrics},
      author={Chae-Yeon, Lee and Hyun-Bin, Oh and EunGi, Han and Sung-Bin, Kim and Nam, Suekyeong and Oh, Tae-Hyun},
      booktitle={Proceedings of the Computer Vision and Pattern Recognition Conference},
      pages={21065--21074},
      year={2025}
    }

@article{ki2024float,
            title={FLOAT: Generative Motion Latent Flow Matching for Audio-driven Talking Portrait},
            author={Ki, Taekyung and Min, Dongchan and Chae, Gyeongsu},
            journal={arXiv preprint arXiv:2412.01064},
            year={2024}
}

@misc{dagan,
            title={Depth-Aware Generative Adversarial Network for Talking Head Video Generation},
            author={Hong, Fa-Ting and Zhang, Longhao and Shen, Li and Xu, Dan},
            journal={IEEE/CVF Conference on Computer Vision and Pattern Recognition (CVPR)},
            year={2022}
}

@InProceedings{fom,
  author={Siarohin, Aliaksandr and Lathuilière, Stéphane and Tulyakov, Sergey and Ricci, Elisa and Sebe, Nicu},
  title={First Order Motion Model for Image Animation},
  booktitle = {Conference on Neural Information Processing Systems (NeurIPS)},
  month = {December},
  year = {2019}
}

@article{wierzbicka2000semantics,
  title={The semantics of human facial expressions},
  author={Wierzbicka, Anna},
  journal={Pragmatics \& cognition},
  volume={8},
  number={1},
  pages={147--183},
  year={2000},
  publisher={John Benjamins}
}
}

\clearpage
\setcounter{page}{1}
\maketitlesupplementary
\appendix

\section{User study}
\label{sec:appendix_user_study}

To ensure a fair and unbiased comparison between methods, the algorithm selects videos randomly from common videos of all methods. By choosing videos at random, the evaluation avoids over-representing any specific content or scenario, which could otherwise skew results. Additionally, the algorithm randomly assigns each method to the left or right position in the user interface for each comparison to mitigates positional bias, ensuring that participants preferences are based on the video quality itself rather than the side on which it appears.
Screenshot of the website is available on Figure~\ref{fig:website}. Each human rater has seen a total of 153 videos for all pairs of videos comprising the TH methods but also the real videos. We use Krippendorff’s $\alpha$ to measures agreement across annotators. Using this metric on all 153 method-pair comparisons, we obtain $\alpha$ = 0.74, indicating substantial agreement among participants.

\section{Additional experiments on Syncnet instability}
\label{Syncnet_exp}
Our experiments showed that Syncnet LSE-C and LSE-D can be influenced by the way audio and video are encoded. Indeed, when changing the audio encoding from \textbf{mp4a} to \textbf{mpga}, the LSE-D and LSE-C vary significantly. When tested on the entire HDTF dataset, we notice that the average absolute difference in LSE-D and LSE-C between videos with \textbf{mp4a} or \textbf{mpga} audio is 0.4. This absolute difference can even reach values as high as 1.2 for some samples. We observe similar results when comparing video using \textbf{H.264} and \textbf{H.265} encodings. In both experiments there are no noticeable qualitative differences from a human evaluation standpoint. This confirms the findings of \citep{yaman2024audio} that Syncnet is not stable and can be influenced by various factor unrelated to lips synchronization.

\section{Temporal Drift in OmniAvatar}
\label{app:omni_drift}

\begin{figure}[htbp]
    \centering
    \includegraphics[width=\columnwidth]{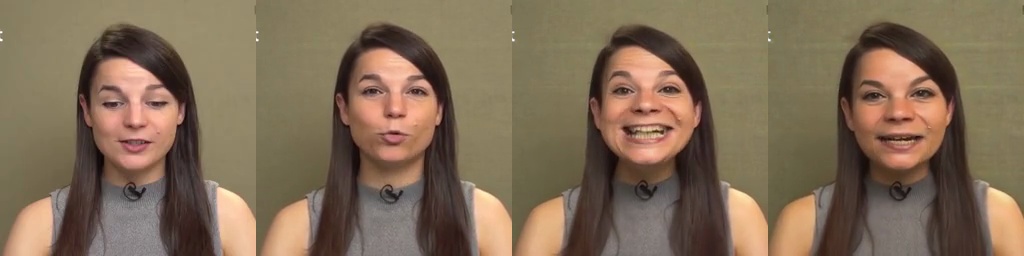}
    \caption{Temporal drift in OmniAvatar outputs over time. Ten frames are shown, sampled every 75 frames, illustrating gradual facial identity divergence, increasing visual artifacts, and the emergence of a color cast in longer videos.}
    \label{fig:omni_d}
\end{figure}

\section{State-of-the-art TH-generation methods used for benchmarking}
\label{methods_desc}
\paragraph{Controltalk} \citep{zhao2024controllable} is a talking face generation method to control face expression deformation based on driven audio, constructing the head pose and facial expression (lip motion) for both single image or sequential video inputs in a unified manner.

\paragraph{DaGAN} \citep{dagan} is a Depth-aware Generative Adversarial Network that first recovers dense 3D facial geometry (i.e., depth maps) from face videos in a self-supervised manner. This depth information is then used to guide the estimation of sparse facial keypoints and to learn a 3D-aware cross-modal attention mechanism, improving the generation of accurate face structures and motion fields.

\paragraph{Dimitra} \citep{chopin2025Dimitra} is a diffusion based framework for TH generation that uses 3D motions as an intermediate step. It leverages audio features, phonemes and text to generate fully animated, realistic TH videos.

\paragraph{EchoMimic} \citep{chen2024echomimic} uses audio speech to drive landmark sequences and employs a Latent Diffusion Model to convert input images into an efficient latent representation that is driven with the landmark sequence. It generates realistic results at high resolution.

\paragraph{EMOPortraits} \citep{drobyshev2024emoportraits} builds upon the MegaPortraits model to enhance its capability for rendering intense and asymmetric facial expressions. It introduces architectural changes and a new training pipeline, including a novel dataset with extreme emotions (FEED), and incorporates a speech-driven mode, making it a multimodal framework for high-fidelity avatar animation.

\paragraph{Hallo2} \citep{cui2024hallo2} generates long-duration, high-resolution audio-driven portrait animations. It uses a patch-drop technique for temporal consistency, vector quantization for high resolution, and supports textual prompts for expression control.

\paragraph{LIA} \citep{wang2022latent} is a self-supervised autoencoder that animates images by linear navigation in its latent space, removing the need for explicit structure representation. Motion is constructed by the linear displacement of latent codes, using a learned set of orthogonal motion directions.

\paragraph{LIA-X} \citep{wang2025lia} is an interpretable portrait animator designed as an autoencoder that models motion transfer as a linear navigation of motion codes. It incorporates a Sparse Motion Dictionary to disentangle facial dynamics into interpretable factors, enabling a controllable 'edit-warp-render' strategy for precise manipulation of facial semantics.

\paragraph{Liveportrait} \citep{guo2024liveportrait} is an efficient video-driven portrait animation framework using an implicit-keypoint-based approach for good generalization and controllability. It features stitching and retargeting modules for precise control over elements including eye and lip movements with minimal computational cost.

\paragraph{MCNet} \citep{mcnet} proposes a Memory Compensation Network to address ambiguities from dramatic motions in talking head generation. It learns a global facial meta-memory bank that provides structure and appearance priors. An implicit identity representation, learned from keypoints and features of the source image, is used to query this memory bank and compensate for warped features, particularly in occluded regions.

\paragraph{FOM} \citep{fom} proposes novel motion representations for animating articulated objects by identifying and tracking object parts as regions rather than keypoints. In a fully unsupervised manner, it infers motion from the principal axes of these regions, disentangles shape and pose to prevent identity leakage, and models global background motion separately with an affine transformation.

\paragraph{OmniAvatar} \citep{gan2025omniavatar} is an audio-driven video generation model focused on creating full-body animations with adaptive and natural movements. It employs a LoRA-based training approach on a foundation model and introduces a multi-hierarchical, pixel-wise audio embedding strategy to enhance lip-sync accuracy and ensure audio features guide the entire body motion, not just the face.

\paragraph{Real3DPortrait} \citep{ye2024real3d} is a framework for realistic 3D talking portrait synthesis. It improves 3D reconstruction by distilling knowledge from a 3D face generative model into an image-to-plane network. It facilitates animation with a motion adapter and synthesizes a complete, realistic video by individually modeling the head, torso, and background, supporting both audio and video-driven inputs.

\paragraph{SadTalker} \citep{zhang2023sadtalker} is a method for generating TH that produces realistic 3D motion coefficients for animated, audio-driven TH from a single image. It leverages full-image animation capabilities and utilizes pre-trained models to enhance the expressiveness and authenticity of the animated TH.

\paragraph{Wav2Lip} \citep{wav2lips} is a lip synchronization model for videos, aligning lip movements with audio segments for different identities in various settings. It uses a lip-sync discriminator based on Syncnet to enhance the precision of lip movements in TH videos. Wav2Lips does not generate the entire TH but only the mouth region. The generated mouth region is then integrated into the original video without altering the rest of the content.

\paragraph{X-Portrait} \citep{xie2024x} is a conditional diffusion model for expressive portrait animation. It uses a pre-trained Stable Diffusion model as a rendering backbone and achieves fine-grained motion control via ControlNet. It interprets dynamics directly from the raw driving video (implicit control) rather than relying on intermediate representations such as landmarks, and uses a cross-identity training scheme to mitigate identity leakage from the driver.

\section{THEval Dataset}
\label{app:dataset_details}

\begin{figure*}[t!]
    \centering
    \includegraphics[width=\textwidth]{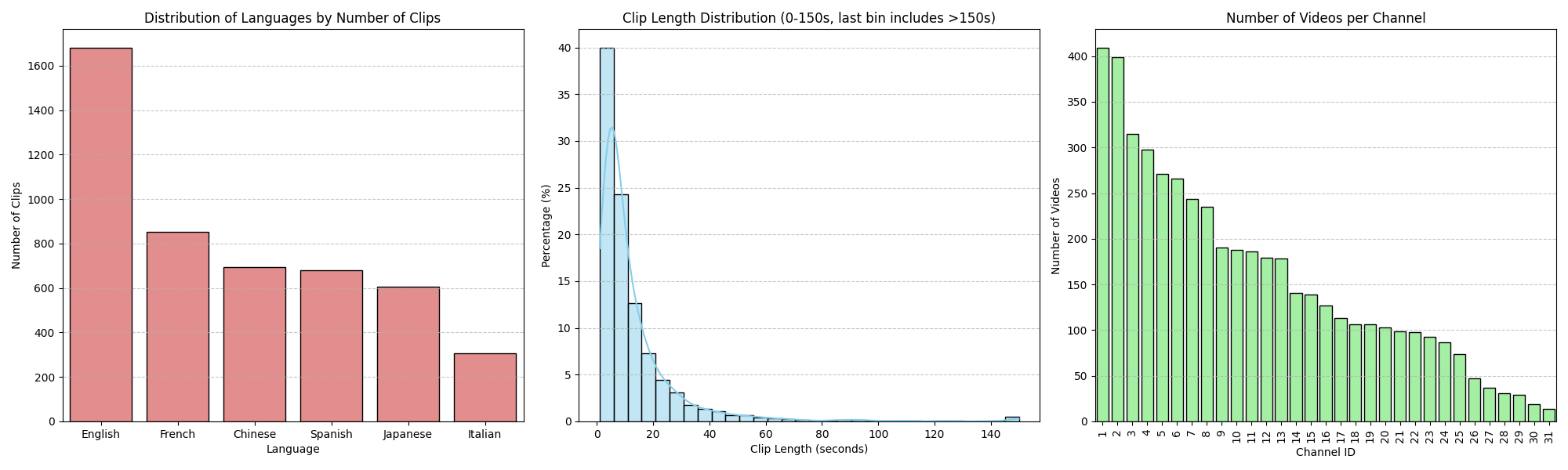}
    \caption{\textbf{Dataset Distribution Details.} (Left) Distribution of the 5,011 clips by language. (Center) Distribution of clip lengths, showcasing a high concentration of short clips. (Right) Distribution of videos across the 31 unique source channels, illustrating channel diversity.}
    \label{fig:combined_distribution}
\end{figure*}

We curate a dataset comprising of TH-videos that includes multiple languages, varied clip lengths, which we have collected from 31 video channels
comprising diverse speakers and settings such as 
 lighting conditions, background, and camera angles. Figure \ref{fig:combined_distribution} illustrates details on the dataset composition. 

The dataset contains a total of {5,011 clips} across {six different languages} to test the model's robustness and generalization capability beyond English-centric content.
In particular, {English} is the most represented language with {1,680 clips}, {French} is included in 890 clips, {Chinese} in 733, {Spanish} in 719, {Japanese} in 645, and {Italian} in 344 clips.


The dataset features a wide range of clip durations, from 2 seconds to 334.6 seconds, with
the majority of clips being short. The first bin in (Center) Fig \ref{fig:combined_distribution} corresponds to (0-10 seconds), accounting for a large percentage of the total clips, which is a typical for many TH datasets. 



\section{Existing metrics for evaluation of TH-generation}
\label{glossary}
In this section we elaborate on the existing metrics for video evaluation, as well as on their limitations.

\textbf{FID:}
The Fréchet Inception Distance (FID) constitutes an improvement of the Inception Score (IS).
is a metric designed to evaluate the quality of generated images or videos. FID is computed by first extracting features from real and generated images using an inception network. Then, the features are treated as samples from two multivariate Gaussian distributions (real and generated) and the Fréchet distance between the two distributions is computed. 

The Fr\'echet distance measures the distance between a generated image set and a source dataset, and is calculated as
\begin{equation}
\text{FID} = || \mu_r - \mu_g ||_2^2 + \text{Tr}( \Sigma_r + \Sigma_g - 2(\Sigma_r \Sigma_g)^{1/2} )
,
\end{equation}
where $\mu_r$ and $\mu_g$ are the mean feature vectors of the real and generated images respectively,  $\Sigma_r$ and  $\Sigma_g$ are the covariance matrices of the real and generated images respectively.

FID is highly dependent on the performance of inception network and assumes that the images features follow a Gaussian distribution which might not be true. FID is also biased when evaluated on a finite set due to the assumption of Gaussian distribution  \citep{Chong_2020_CVPR}. To be accurate, FID must be evaluated on a set that is large enough which might not be possible for all generation tasks.
When used for video evaluation FID will only evaluate independent frame quality without regards for the temporal coherency.

\textbf{FVD:}
The Fréchet Video Distance (FVD) is similar to FID but uses a network adapted for videos to extract the features.
FVD is calculated as
\begin{equation}
\text{FVD} = || \mu_r - \mu_g ||_2^2 + \text{Tr}( \Sigma_r + \Sigma_g - 2(\Sigma_r \Sigma_g)^{1/2},
\end{equation}
where $\mu_r$ and $\mu_g$ are the mean feature vectors of the real and generated videos respectively,  $\Sigma_r$ and  $\Sigma_g$ are the covariance matrices of the real and generated videos respectively.

FVD has the same limitations as FID, and despite using an adapted network FVD still tends to focus more on single frame quality than on temporal coherence which is essential to evaluate videos.

\textbf{IS:}
The Inception Distance (IS), uses an inception network that gives the probability of an image to belong to a certain class. Then, it uses the Kullback-Leibler divergence to compute a score related to the quality and diversity of the generated images. Specifically, the score is calculated to evaluate two factor: Intra-Class Similarity (high-quality images should have a strong probability of belonging to a single class) and inter-Class Diversity (generated images should belong to a variety of classes) and is calculated as
\begin{equation}
\text{IS} = \exp \left( \mathbb{E}_{x \sim p_g} \left[ \text{KL}( p(y|x) || p(y) ) \right] \right),
\end{equation}
Where $x \sim p_g$ denotes that $x$ is sampled from the generated images distribution $p_g$, $p(y|x)$ is the conditional probability distribution over the classes given the image $x$ and $p(y)$ is the marginal probability distribution over the classes, computed as $\mathbb{E}_{x \sim p_g}[p(y|x)]$.

However similarly to FID, this metric is very reliant on the inception network. It is unable to evaluate the intra-class diversity and will not work with images of classes not seen during the training of the inception network.

\textbf{LMD-F:}
The Landmark Distance Face (LMD-F) is a metric to evaluate TH videos. LMD-F computes the average euclidean distance between the facial landmarks extracted for a real videos and those of a generated one for the same conditioning input (e.g driving audio speech). For LMD-F all of the face landmarks are used. LMD-F is calculated as
\begin{equation}
\text{LMD-F} = || x_f - x'_f ||_2
\end{equation}
where $x_f$ and $x'_f$ are the facial landmarks of the real and generated video respectively.

While LMD-F has been shown to correlate better than other metrics with human evaluation \citep{10647543}, it still suffers from being a direct comparison to the ground truth. Indeed, different head motions and expressions between the generated sequence and the ground truth will be strongly penalized. At the same time, this difference is expected since head motion and expression are only loosely correlated to the audio sequence. However, as long as both look natural, human evaluators will give a high rating to the video even if it is different from the ground truth. 

\textbf{LMD-M:}
The Landmark Distance Mouth (LMD-M) is a metrics to evaluate TH videos. LMD-M compute the average euclidean distance between the facial landmarks extracted for a real videos and those of a generated one for the same conditioning input (e.g driving audio speech). For LMD-M only the landmarks pertaining to the mouth area landmarks are used.
\begin{equation}
\text{LMD-M} = || x_m - x'_m ||_2
\end{equation}
where $x_m$ and $x'_m$ are the mouth landmarks of the real and generated video respectively.

While LMD-M has been shown to correlate better than other metrics with human evaluation \citep{10647543}, it still suffers from being a direct comparison to the ground truth. This direct comparison causes small temporal lags to be penalized when it wouldn't be noticed by human evaluators according to the recommendation by the International Telecommunication Union  \citep{1998RecIB}.

\textbf{LPIPS:}
The Learned Perceptual Image Patch Similarity (LPIPS) measures the perceptual similarity between two images and try to provide a score that align with human perception. LPIPS uses a pre-trained CNN to obtain deep-features and computes the similarity between these features. The LPIPS value of CNN layer $l$ is calculated as
\begin{equation}
\text{LPIPS}_l(x, x') = \sum_{l} w_l \cdot || f_l(x) - f_l(x') ||_2,
\end{equation}
where $f_l(x)$ and $f_l(x')$ are the feature representations of the real image $x$ and the generated image $x'$ at layer $l$, $w_l$ are the weights of layer $l$. The final LPIPS score is a weighted sum of the $\text{LPIPS}_l$ across all the layers of the network.

While LPIPS aligns better with human evaluation it is still very dependent on the pre-trained network and is sensitive to image alignment.

\textbf{PSNR:}
The Peak Signal-to-Noise Ratio (PSNR) compares two images at the pixel level by measuring the ratio between the maximum possible power of a signal (the original image) and the power of corrupting noise (the generated image). It is calculated as
\begin{equation}
\text{PSNR} = 10 \cdot \log_{10} \left( \frac{{\text{MAX}^2}}{{\text{MSE}}} \right)
\end{equation}
\begin{equation}
\text{MSE} = \frac{1}{mn} \sum_{i=1}^{m} \sum_{j=1}^{n} \left[ I(i,j) - K(i,j) \right]^2,
\end{equation}
where $\text{MAX}$ is the maximum possible pixel value of the image (usually 255), $I(i,j)$ represents the pixel value at position $(i,j)$ in the original image, and $K(i,j)$ represents the pixel value at position $(i,j)$ in the reconstructed image. The sums are taken over all pixels in the $m \times n$ image.

PSNR does not take the structure of the image into account, is very sensitive to noise and to outliers which can lead to low correlation with human evaluation.

\textbf{SSIM:}
The Structural Similarity (SSIM) is a score that evaluates the similarity between two images. It is obtained by combining three components : the difference in brightness between the images, the difference in contrast between the images and the structural similarities between the images across small patches. SSIM is calculated as
\begin{equation}
    \text{SSIM}(x, y) =  l(x, y) \cdot c(x, y) \cdot s(x, y)
\end{equation}
\begin{equation}
l(x, y) = \frac{2 \mu_x \mu_y + C_1}{\mu_x^2 + \mu_y^2 + C_1}
\end{equation}
\begin{equation}
c(x, y) = \frac{2 \sigma_x \sigma_y + C_2}{\sigma_x^2 + \sigma_y^2 + C_2}
\end{equation}
\begin{equation}
s(x, y) = \frac{\sigma_{xy} + C_3}{\sigma_x \sigma_y + C_3},
\end{equation}
where $\mu_x$ and $\mu_y$ are the means of the images $x$ and $y$, $\sigma_x$ and $\sigma_y$ are the standard deviations of the images $x$ and $y$, $\sigma_{xy}$ is the covariance between the images $x$ and $y$, $C_1$, $C_2$, and $C_3$ are small constants to stabilize the division when the denominators are close to zero. $l(x, y)$, $c(x, y)$ and $s(x, y)$ correspond to the luminance comparison, contrast comparison and structure comparison respectively.

SSIM need the images to be perfectly aligned in order to be accurate. Also, since it use small patches, it focus on local structure rather than global which lead to low correlation with human perception on complex images.

\balance

\textbf{Syncnet:}
Syncnet \citep{Chung16a} is a CNN-based network, aims to capture the correlation between audio and spatio-temporal features of the mouth region, calculating the audio offset (the number of frames by which audio is early or late compared to video). Its distance (LSE-D) and confidence score (LSE-C) are widely use to evaluate audio-lip synchronization in TH video. While Syncnet is good at evaluating the audio offset, finding the speaker in a video containing multiple persons or detecting unrelated audio (e.g dubbing) it is less useful when comparing two videos with similar lip synchronization (e.g videos generated by two different methods). In fact it has been shown that LSE-C and LSE-D have very limited correlation with human evaluation \citep{10647543}. Some recent methods \citep{xu2024vasa} were even able to outperform the ground truth by a large margin on these metrics, showcasing their limitations. Additionally recent works \citep{yaman2024audio,yaman2024audio_2} have shown that Syncnet is not stable and can easily be influenced by factors outside of lip synchronization (e.g mouth cropping, image quality, brightness...) making it difficult to apply on the diverse datasets used today. Additionally, our own experiments have shown that Syncnet is sensitive to audio and video encoding even when there are not noticeable difference for a human observer (SM~\ref{Syncnet_exp})

\textbf{LSE-D:}
The Syncnet Distance (LSE-D) compute the distance  between audio and video features at the offset predicted by Syncnet. See \textbf{Syncnet} entry for limitations. 

\textbf{LSE-C:}
The Syncnet Confidence score (LSE-C) computes the difference between the minimum and the median of the features distances over all possible offsets (\( -10 \leq offset \leq 10, \quad offset\in \mathbb{Z} \)). See \textbf{Syncnet} entry for limitations. 



\end{document}